# Driving Tasks Transfer in Deep Reinforcement Learning for Decision-making of Autonomous Vehicles

Hong Shu, Teng Liu, *Member*, *IEEE*, Xingyu Mu, Dongpu Cao

*Abstract*—Knowledge transfer is a promising concept to achieve real-time decision-making for autonomous vehicles. This paper constructs a transfer deep reinforcement learning framework to transform the driving tasks in the intersection environments. The driving missions at the unsignalized intersection are cast into a left turn, right turn, and running straight for automated vehicles. The goal of the autonomous ego vehicle (AEV) is to drive through the intersection situation efficiently and safely. This objective promotes the studied vehicle to increase its speed and avoid crashing other vehicles. The decision-making policy learned from one driving task is transferred and evaluated in another driving mission. Simulation results reveal that the decision-making strategies related to similar tasks are transferable. It indicates that the presented control framework could reduce the time consumption and realize online implementation.

## NOMENCLATURE

| | |
|---|---|
| AEV | Autonomous Ego Vehicle |
| AI | Artificial Intelligence |
| DRL | Deep Reinforcement Learning |
| DP | Dynamic Programming |
| RL | Reinforcement Learning |
| TL | Transfer Learning |
| TRL | Transfer Reinforcement Learning |
| DQL | Deep Q-learning |
| MDP | Markov Decision Process |
| NN | Neural Network |

This work was supported by Chongqing Technology Innovation and Application Development Special Major Theme Special Project, Chongqing Science and Technology Bureau, under Grant cstc2019jscx-zdztzxX0039.

H. Shu and X. Mu are with Department of Automotive Engineering, Chongqing University, Chongqing 400044, China (email: shuhong@cqu.edu.cn, 20162364@ cqu.edu.cn, Corresponding author: H. Shu)

T. Liu is with Department of Automotive Engineering, Chongqing University, Chongqing 400044, China, and also with Department of Mechanical and Mechatronics Engineering, University of Waterloo, N2L 3G1, Canada. (email: tengliu17@ gmail.com)

D. Cao is with Department of Mechanical and Mechatronics Engineering, University of Waterloo, N2L 3G1, Canada. (email: dongpu.cao@uwaterloo.ca)

| | |
|---|---|
| IDM | Intelligent Driver Model |

## I. INTRODUCTION

AUTONOMOUS driving enables the human drivers to have an enjoyable, efficient and safe driving experiences beyond the traditional driving [1]-[2]. In order to achieve self-driving with high automation, many high technologies are necessary. Inspired by the significant development of artificial intelligence (AI), many car manufacturers have constructed their own autonomous vehicles. Deep reinforcement learning (DRL) is regarded as a promising methodology to establish the functional modules for automated vehicles [3]-[4]. However, how to build the real-time decision-making controller is still a challenging topic [5].

Decision-making modules provide a sequential sequence of driving behaviors for autonomous vehicles to accomplish the stationary driving missions [6]-[7]. The typical driving scenarios are highway and urban situations. Including the pedestrians, bicycles and classical motor vehicles as participants, it is especially difficult to realize autonomous driving in urban conditions. Intersection is a representative driving situation, and many references have discussed the self-driving problems at intersections [8]. For example, the authors in [9] constructed a distributed intersection protocol for connected automated cars to access a traffic junction. The controller is based on the potential functions, and the protocol is verified by the in-vehicle experimental test. Miculescu et al. [10] proposed a coordination control algorithm to help the self-driving cars drive through the intersections with no traffic signals. The simulation results indicate that the presented control framework could guarantee the safety and performance. Furthermore, Ref. [11] aims to plan the speed trajectories for connected and automated vehicles at signalized intersections. Dynamic programming (DP) is applied to resolve the optimization problem and the robustness is capable of being improved in the face of random signals.

Reinforcement learning (RL) and DRL are widely used in the decision-making problems for self-driving vehicles [12]. To date, the combination of transfer learning (TL) and RL (or DRL) is a hopeful research concept to formulate the real-time decision-making controller [13]. Some attempts have been conducted to realize the implementations of transfer reinforcement learning (TRL). For example, Hou et al. [14] presented an evolutionary transfer reinforcement learning framework for multiagent systems. The efficacy of this paradigm is demon- strated by comparing with the common TL methods. To address the multi-perspective light field reconstruction problem, the authors in [15] applied TRL idea to learn the feature set of the source domain and target domain.

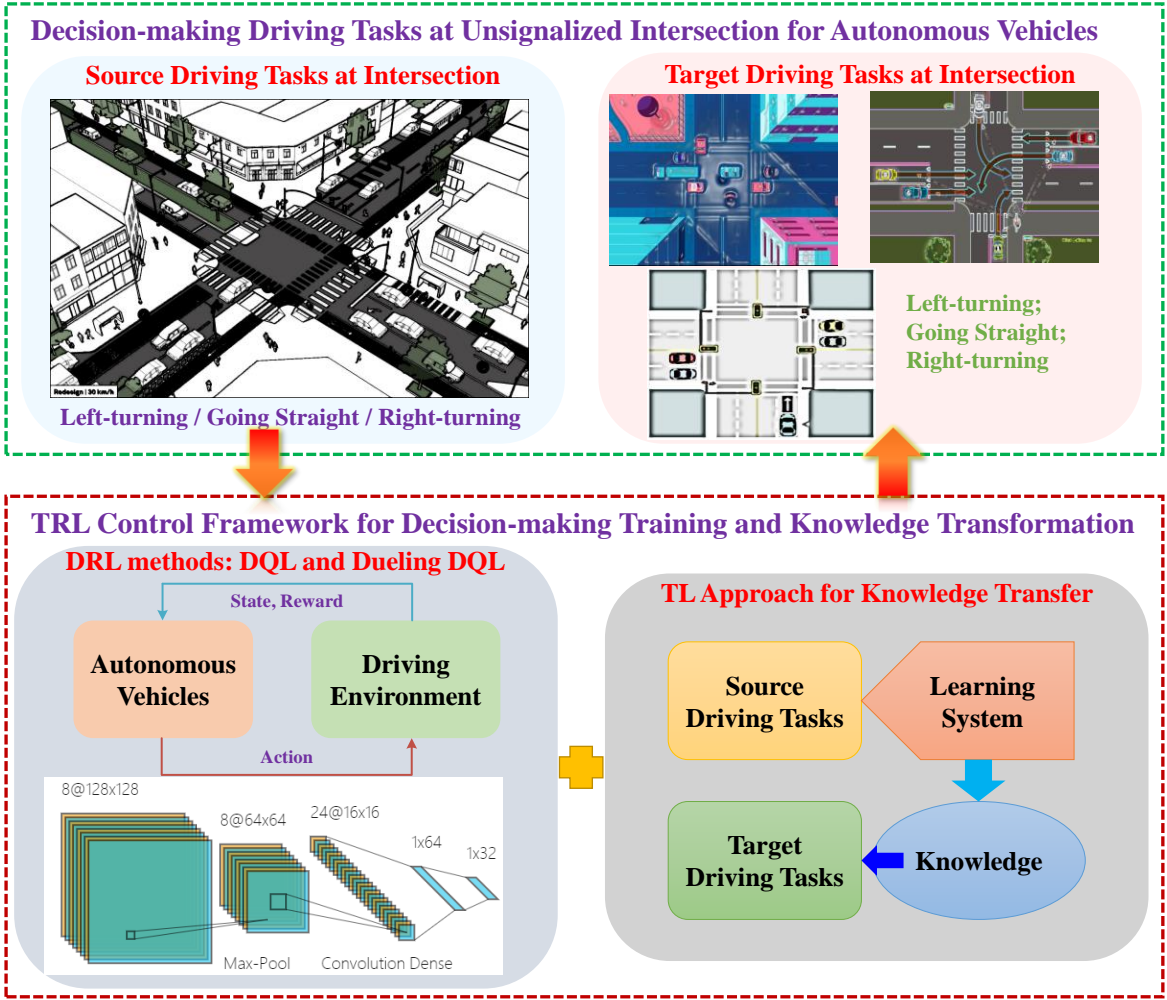

Fig. 1. Control architecture of the presented TRL for decision-making policies transformation for autonomous vehicles.

Furthermore, Ref. [16] and [17] discussed the trajectory planning and target recognition topics for autonomous vehicles using TRL. Simulation results imply that the proposed TRL algorithms are able to ensure the model's generalization and real-time performance.

Motivated by the successful applications of TRL approaches in other research fields, this paper proposes a decision-making control framework at intersections for automated vehicles based on TL and DRL. The schematic diagram of this control structure is depicted in Fig. 1. The DRL technique and TL concept are firstly introduced. The special DRL algorithm is dueling deep Q-learning (DQL), whose performance has been certified in [18]. Then, the driving scenarios of the intersection situations are described. The behaviors controller is given to manipulate the AEV and surrounding vehicles. Finally, multiple simulation experiments have been designed to evaluate the performance of TRL-enabled decision-making policy. The advantages of the proposed decision-making modules are analyzed and demonstrated.

The original innovations and contributions of this work are illuminated as follows: 1) a TRL-enabled control framework is built to resolve the decision-making problems for autonomous vehicles; 2) the dueling DQL algorithm is combined with TL method to transform the learned knowledge of different driving tasks at intersections; 3) a series of reasonable tests have been constructed to evaluate the optimality and real-time performance of the presented decision-making strategies. This work is one attempt to apply the TRL concept to establish the real-time decision-making policy for autonomous vehicles.

To better explain the content of this paper, the rest of this article is arranged as follows. Section II describes the RL framework, dueling DQL algorithm and TL idea. The driving scenarios and behavior controller are depicted in Section III. Section IV discusses the related experiment results of the constructed decision-making policy. Finally, the concluding remarks are provided in Section V.

## II. TRANSFER DEEP REINFORCEMENT LEARNING

In this section, the transfer deep reinforcement learning method is introduced. First, the parameters of RL framework is built by Markov decision process (MDP). The realization of the dueling DQL algorithm is then given. The merits of this algorithm and its applicable scenes are analyzed. Furthermore, the TL is utilized to transfer the trained knowledge of different driving tasks.

### A. RL Concept

In RL framework, an intelligent agent aims to search the optimal control sequence. The selection process is guided by the environment. By interacting with the environment, the agent could improve the quality of the control action. This intersection is always mimicked by a finite Markov decision processes (MDPs), as a tuple $(\mathcal{S}, \mathcal{A}, \mathcal{P}, \mathcal{R})$. $\mathcal{S}$ and $\mathcal{A}$ represent the state space and action space. $\mathcal{R}$ is the reward function, which is determined by the state variable and control action. Finally, $\mathcal{P}$ is the transition model to indicate the transformation of state variable [19].

RL methods have been applied in many research domains to derive the optimal control strategies [20-24]. In these problems, the goal of the agent is maximizing the accumulated reward. This reward is the sum of the immediate reward and discounted future rewards as follows:

$$R_t = \sum_{k=t}^{T} \gamma^{k-t} \cdot r_k \tag{1}$$

where $t$ is the time step, and $T$ is the time limitation. $r_k$ is the instantaneous reward received from the environment. $\gamma \in [0, 1]$ is a discount parameter to realize the trade-off of current and future rewards.

The control policy obtained by the agent is $\pi$. To compute the control action at each time instant easily, the cumulative reward is expressed as the expected forms, which are named as value functions. Two value functions are represented via the relevant state variable $s_t$ and control action $a_t$ as:

$$V_\pi(s_t) \doteq E_\pi \left[ \sum_{k=t}^{T} \gamma^{k-t} \cdot r_k \mid s_t \right] \tag{2}$$

$$Q_\pi(s_t, a_t) \doteq E_\pi \left[ \sum_{k=t}^{T} \gamma^{k-t} \cdot r_k \mid s_t, a_t \right] \tag{3}$$

where (2) and (3) are the state-value function and action-value function. $E_\pi [\cdot]$ denotes the expected value of the accumulated reward following policy $\pi$. The significance of one RL algorithm is to update these value functions. Furthermore, many RL algorithms are motivated to update the action-value function (Q-table for short), wherein the action $a_t$ is included. Depend on the disparate renewed criterions of Q-table, RL algorithms are classified into multiple categories. They are model-based and model-free, Monte Carlo and temporal difference, value-based and policy-based, and on-policy and off-policy [25].

To conveniently generate the control action at each step, the action-value function is rewritten as the recursive form:

$$Q_\pi(s_t, a_t) = E_\pi \left[ r_t + \gamma \max_{a_{t+1}} Q_\pi(s_{t+1}, a_{t+1}) \right] \tag{4}$$

where $s_{t+1}$ and $a_{t+1}$ are the state and action at the next step. The optimal control action is interpreted as the action maximizing the action-value function. The Bellman optimality equation is used to compute the maximum value of Q-table:

$$Q^*(s_t, a_t) = \sum_{s_{t+1}, r_t} p(s_{t+1}, r_t \mid s_t, a_t) \left[ r_t + \gamma \max_{a_{t+1}} Q^*(s_{t+1}, a_{t+1}) \right] \tag{5}$$

where $p(s_{t+1}, r_t \mid s_t, a_t) \in \mathcal{P}$ is the transition probability in transition model. The tuple $(s_{t+1}, r_t, s_t, a_t)$ is named as a transition or an observation in the RL algorithm to record the intersection of the agent and environment. Finally, the optimal control action is derived from the maximum Q-table as follows:

$$a_t^* = \arg \max_{a_t} Q(s_t, a_t) = \arg Q^*(s_t, a_t) \tag{6}$$

Having ostensive action in the expression, many RL algorithms are defined to construct a mature action-value function (instead of state-value function). In this work, the conventional DQL and dueling DQL algorithms are introduced and exploited to update the Q-table. These algorithms are displayed in the next subsection.

### B. Dueling DQL Algorithm

With the enormous ability to recognize the underlying relationships in a set of data, neural network (NN) is a powerful tool to build connections between the inputs and outputs. Deep reinforcement learning (DRL) is a combination of deep learning (or the NN) and RL [26]. The classical DRL algorithm is the deep Q-learning (DQL), which is first used to play the Atari games [27]. Reformulating the observation in the RL algorithm as $(s', r, s, a)$, the Q-learning algorithm is described as the following shape to renew the Q-table:

$$Q(s, a) \leftarrow Q(s, a) + \alpha[r + \gamma \max_{a'} Q(s', a') - Q(s, a)] \tag{7}$$

where $\alpha \in [0, 1]$ is the learning rate in RL methods. It is applied to balance the new and old collected experiences (or transitions) from the environment. In the RL algorithms, the epsilon greedy strategy is utilized to choose the control action from action space at each step. This policy propels the agent to explore a random action with probability $\varepsilon$, and to exploit the current best action with probability 1-$\varepsilon$.

For a complex problem with large state space and action space, the Q-table would be an immense matrix. It is time-consuming to obtain a mature Q-table. Hence, in DQL, the neural network is employed to approximate the Q-table as Q(s, a; $\omega$), wherein $\omega$ represents the weights and biases of NN. The inputs of the NN are the state variables and control actions, and the outputs are the estimated Q-table.

Two techniques are often used to update the Q-table in DQL, which are fixed target networks and experience replay. The former means the evaluate and target networks exist simultaneously in the training process of NN. The target network is a transcript of evaluate network, and it would be altered with a special step. The latter indicates that a series of transitions (or observations) are stored in a finite-sized cyclic buffer. A mini-batch of these samples is randomly chosen to train the evaluate network instead of the current experiences [28]. To measure the differences between the trained network and actual Q-table, the loss function is introduced as:

$$L(\omega) = E \left[ \left( r + \gamma \max_{a'} Q(s', a'; \omega') - Q(s, a; \omega) \right)^2 \right] \tag{8}$$

where $\omega'$ and $\omega$ are the parameters of target network and evaluate network, respectively. By updating the target network with the parameters of evaluate network periodically, the training performance is guaranteed to be promoted steadily.

Signing $y = r + \gamma \max_a Q(s', a'; \omega')$, the gradient descent method is usually applied to update the evaluate network as follows:

$$\nabla_\omega L(\omega) = E\left[\left((y - Q(s,a;\omega))\nabla_\omega Q(s,a;\omega)\right)\right] \quad (9)$$

For some optimization control problems, the current control action may not cause apparent negative effects for the agent. It implies that the states have no repercussion with the instant action. The decision-making problem at the intersection of this work is one of these examples. Regarding to this kind of problem, dueling DQL is demonstrated to be more effective than the conventional DQL [29].

In dueling DQL algorithm, a dueling network is established to approximate the action-value function (Q-table). Its essence of this network is a combination of two neural networks. One NN is used to calculate the state-value function $V(s)$, and another one aims to compute the advantage function $A(s, a)$ [29]. This advantage function is the main innovation of the dueling DQL. It represents the consequences of control actions on the state variables. For example, the lane-changing behavior may not cause a collision immediately for self-driving vehicle. However, it would urge the objective vehicle to crash other vehicles soon afterwards. The sketch of the dueling network is depicted in Fig. 2.

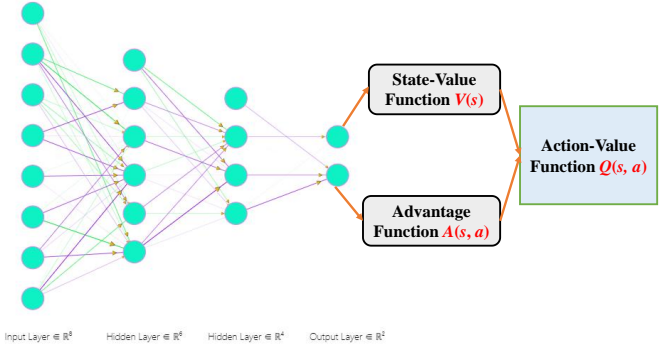

Fig. 2. Dueling network for action-value function approximation.

In this work, the action-value function in dueling DQL is written as follows:

$$Q(s,a;\omega,\varsigma,\tau) = V(s;\omega,\varsigma) + A(s,a;\omega,\tau) \quad (10)$$

where $\tau$ and $\varsigma$ are the parameters of NN to approximate the value-action function and advantage function. For generation of the optimal control action using dueling DQL, (10) is formulated as forward mapping:

$$\begin{aligned}
Q(s,a;\omega,\varsigma,\tau) &= V(s;\omega,\varsigma) + A(s,a;\omega,\tau) \\
&= V(s;\omega,\varsigma) + \left(\begin{array}{c} A(s,a;\omega,\tau) - \\ \dfrac{1}{|\mathrm{A}|}\sum_{a'} A(s,a';\omega,\tau) \end{array}\right)
\end{aligned} \quad (11)$$

where $|\mathcal{A}|$ is the norm of the action space, $a'$ is the next action. The optimal action is derived by (11) as the following expression:

$$a^* = \arg\max_{a'} Q(s,a';\omega,\varsigma,\tau) = \arg\max_{a'} A(s,a';\omega,\tau) \quad (12)$$

In this article, the dueling DQL is incorporated with TL idea to transfer the learned knowledge in the decision-making problem for autonomous vehicles. The dueling network is trained by the gradient descent technique in (9). The trained parameters of these networks from one driving task are transformed and applied into another driving mission under TL framework. By doing this, the training time can be reduced sharply. Taking the control performance as the primary goal, this transfer reinforcement learning (TRL) concept is capable of being applied in online implementations.

### C. TL for Knowledge Transformation

The selection of control actions by the intelligent agent in RL is a trial and error process. It is always time-consuming. For different studied problems, the training progress needs to be repeated. Moreover, the performance of DRL methods is extremely dependent on the setting of hyperparameters. How to improve the training efficiency and ensure the control effectiveness is still a challenge in DRL research.

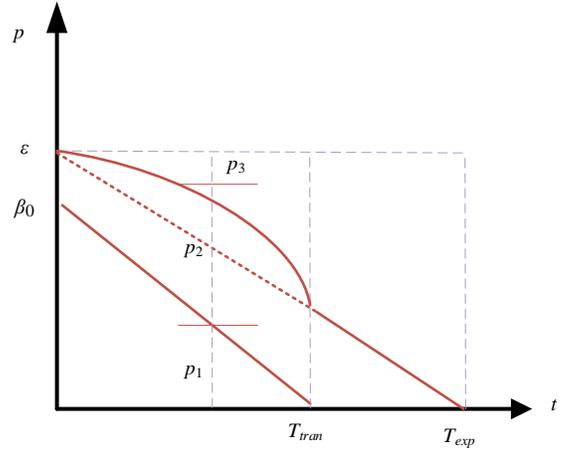

Fig. 3. Changed epsilon greedy policy for action selection in target task.

Transfer learning (TL) technique is a novel paradigm of machine learning [30]. This approach tries to transform the learned knowledge from the source tasks to the target tasks. For example, regarding to the studied decision-making problem in this paper, the goal is to drive as fast as possible without collision. The related tasks are turning left, going straight and turning right at the unsignalized intersection. Hence, the control policy derived from one task is able to be applied in another mission. The integration of TL and DRL is an efficient modality to promote the learning efficiency. Several attempts have been made in different research fields [31, 32].

For the DRL methods, the transformation in TL means the reuse of the parameters of NN. The NN in the source tasks is named expert network, and the NN in the target tasks is the student network. For specification in this work, the TL represents altering the epsilon greedy strategy for control action choice. Three rules are defined to determine the control action in the target tasks [33], as shown in Fig. 3. The **transfer rule** states that with probability $p_1$, the agent chooses the action suggested by the expert network. The related probability is described as follows:

$$p_1 = \beta_0 \left(1 - t / T_{tran}\right) \tag{13}$$

where $\beta_0$ is the initial transfer belief, and $T_{tran}$ is the transfer period, during which the agent in the target network is affected by the expert network. $T_{exp}$ is the initial exploration period in the target task.

The second rule is **exploration**, which implies that the agent selects a random action with probability $p_2 = \varepsilon(1-p_1)$. The third rule is **exploitation**, and it indicates that the agent chooses the best control from the student network with probability $p_3 = (1 - \varepsilon)(1 - p_1)$. Guiding by these three rules, the agent could efficiently utilize the knowledge from the expert network. This feature is capable of maintaining the performance and elevate the training efficiency in the target tasks.

In this work, the driving tasks are three types, which are turning left, turning right and going straight. For the automated vehicle in the intersection environment, one of these missions is set as a source task, and the remaining two tasks are target tasks. For different source tasks, we evaluate the learned expert in the above three tasks. The relevant drivability, collision conditions, and learning efficiency of the studied autonomous vehicles are estimated. In the next section, the driving scenario, behavior controller and default parameters of the DRL technique are expounded, wherein the TRL method is employed to construct the decision-making strategy for the self-driving vehicles.

## III. Vehicles in Intersection Environment

In this section, the unsignalized intersection situation is described. First, the traffic environment is constructed, wherein the autonomous vehicle and its surrounding vehicles are located in the four-way intersection. Then, the behavior controller of the surrounding vehicles is built. The vehicle speed and position of these vehicles are randomly settled. Finally, the default parameters of the decision-making are illustrated, including the state variables, control actions, reward function and transition model.

### A. Traffic Environment at Intersection

When an AEV approaches the intersection, it must decide whether to continue along the planned route or to stop in front of the intersection. The decision that whether it should turn to cross the intersection or whether other vehicles have priority depends on the right-of-way. Priority-controlled intersections and right-hand priority intersections are common unsignalized intersection types [34]. This paper selects the priority-controlled intersections as the intersection type, considering that the horizontal road is the main road and the vertical road is the secondary road. At priority-controlled intersections, drivers from the secondary road must give way to drivers from the main road.

This paper mainly studies the decision-making strategy of AEV under different driving tasks. The road condition of intersection is complex, AEV may have potential conflicts with surrounding vehicles, which will lead to stopping or deceleration, and bring certain safety risks. The ideal goal of AEV is to reach the destination as soon as possible without collision.

Therefore, efficiency and safety are selected as the evaluation criteria for left-turn decision-making at intersections, and reflected in the reward function.

As shown in Fig. 4, the possible actions of the vehicles at intersection include turning right, turning left and going straight, and the driving intention of the surrounding vehicles is unknown to AEV, so it is necessary for AEV to find out the driving intention and make the proper decision. In addition, since the location, speed, destination and driving behavior of the surrounding vehicles are generated randomly, it may occur that one vehicle collides with another. To prevent this from happening, a rudimentary collision prediction was added to the behavior of surrounding vehicles. Finally, each surrounding vehicle can predict the future position of other surrounding vehicles in three seconds, so that the surrounding vehicles can avoid collision with each other.

Considering the general situation of the intersection, set the parameters of unsignalized intersection as follows: the frequency of simulation is 20 Hz, and the duration is 15 seconds. The number of vehicles in the surrounding environment is $N = 15$. The position of start and end of AEV is the same in each trial, and AEV makes action selection once every second.

### B. Behavior Controller

In this paper, a double-layer framework is used to control the behavior of surrounding vehicles and AEV. The upper frame manages the longitudinal behavior and the lower frame manages the lateral behavior. In detail, for AEV, the upper framework uses decision-making based on DRL to manage longitudinal behavior, while the lower framework implements a low-level controller which allows AEV to track a given target speed and follow a target lane. For the surrounding vehicles, the lower frame is the same as AEV, using a low-level controller to manage the lateral behavior, while the upper frame utilizes an intelligent driver model (IDM).

More concretely, the lateral controller is a simple proportional-derivative controller, combined with some non-linearities. The lateral controller consists of two parts, position control and heading control. The expression of heading control is as follows:

$$\psi_t = \psi_L + \Delta\psi_t \tag{14}$$

$$\dot{\psi}_t = K_{p,\psi}(\psi_t - \psi) \tag{15}$$

$$\delta = \arcsin(\frac{1}{2}\frac{l}{v}\dot{\psi}_t) \tag{16}$$

where $\psi_t$ is the target heading to follow the lane heading and position, $\psi_L$ is the lane heading and $K_{p,\psi}$ is the heading control gain. $\psi$ is the heading angle, $\delta$ is the steering angle, $v$ is the forward speed, and $l$ is the vehicle length. Position control is represented as follows:

$$v_l = -K_{p,l}\Delta d_l \tag{17}$$

$$\Delta\psi_t = \arcsin(\frac{v_L}{v}) \tag{18}$$

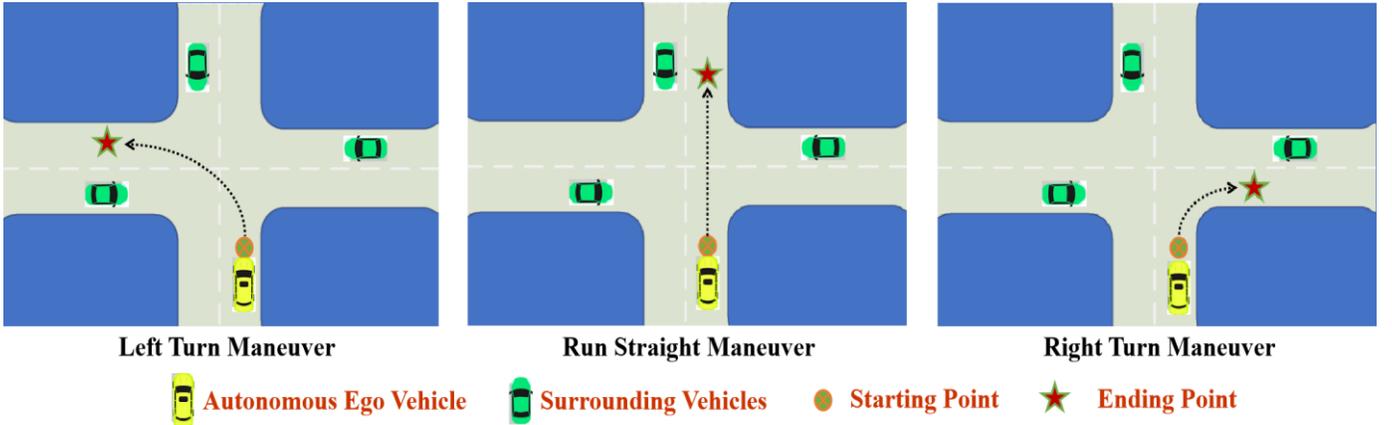

**Left Turn Maneuver**  **Run Straight Maneuver**  **Right Turn Maneuver**

🟨 **Autonomous Ego Vehicle**  🟩 **Surrounding Vehicles**  🔴 **Starting Point**  ★ **Ending Point**

Fig. 4. Decision-making problems at the unsignaled intersection for autonomous vehicles.

where $v_l$ is the required lateral velocity, $K_{p,l}$ represents the position control gain and $\Delta d_l$ is the lateral position of the vehicle with respect to the lane center-line.

For surrounding vehicles, the upper framework uses IDM to manage longitudinal behavior, which helps generate realistic features of road conditions [35]. The desired distance $d_d$ can be calculated by the following formula:

$$d_d = d_0 + T \cdot v + \frac{v \cdot \Delta v}{2\sqrt{a_{max} \cdot b_{max}}} \tag{19}$$

where $d_0$ is the minimum distance between one vehicle and another on the same lane, and $T$ is the desired time gap. $a_{max}$ is the maximum acceleration and $b_{max}$ is the maximum deceleration according to the comfortable purpose. $\triangle v$ is the relative velocity difference between the EAV and its front one, the acceleration of the vehicle $a$ is obtained by equation (20):

$$a = a_{max} \cdot \left(1 - \left(\frac{v}{v_d}\right)^\lambda - \left(\frac{d_d}{d}\right)^2\right) \tag{20}$$

where $v_d$ is the desired velocity, and $\lambda$ is the constant exponent of the velocity term. $d$ is the distance between the leading vehicle and AEV, and $d_d$ is the desired distance. The main parameters in IDM are shown in Table I.

TABLE I
MAIN PARAMETERS IN IDM

| Keyword | Value | Unit |
| --- | --- | --- |
| Comfortable acceleration $a_{max}$ | 6 | m/s² |
| Acceleration argument $\lambda$ | 4 | / |
| Desired time gap $T$ | 1.5 | s |
| Comfortable deceleration $b_{max}$ | -3 | m/s² |
| Minimum relative distance $d_0$ | 7 | m |

### C. Default Setup of Parameters

As described in section II, this paper regards the decision-making of intersection as an MDP, which is represented by state space, action space, a transition model, and a reward function. These settings are described below in this part.

For state-space, the focus of observation is the position $s$ and speed $v$ of the vehicle. In order to facilitate the calculation, the position and speed are respectively expressed in the horizontal direction and vertical direction of the intersection. So for each vehicle observed, its state $S_i = \{s^i_x, s^i_y, v^i_x, v^i_y\}$, where i means the index of the vehicle, and $x$ represents the horizontal direction of the intersection, and $y$ represents the vertical direction of the intersection. Finally, the state space can be expressed as $S = \{S_a, S_1, S_2, \ldots, S_n\}$, where $a$ means the index of the AEV, and $n$ is the total number of surrounding vehicles observed.

As mentioned above in section II, the steering angle $\delta_t$ of the AEV is controlled by a low-level lateral controller. Therefore, the selection of action is decided by DRL and only affects the longitudinal acceleration of the AEV. The action space is defined as follow:

$$a_t \in [-5, \ 0, \ 5] \ \text{m/s}^2 \tag{21}$$

The transition model is established by using the 2-DOF bicycle model to get the vehicle speed and position after state transition. The velocity of the vehicle is easily obtained by acceleration, as shown in equation (22):

$$v_{t+1} = v_t + a_t \cdot dt \tag{22}$$

where $v_{t+1}$ represents the vehicle velocity of next time step after $v_t$, and $dt$ is the length of one time-step. The heading angle $\psi_{t+1}$ of the vehicle can be calculated from the following formulas:

$$\beta_t = \arctan(1 / 2 \tan \delta_t) \tag{23}$$

$$\omega_t = 2v_t \cdot \sin \beta_t \ / \ l \tag{24}$$

$$\psi_{t+1} = \psi_t + \omega_t \cdot dt \tag{25}$$

where $\delta_t$ is the front wheel angle, and $\omega_t$ is the yaw rate of the vehicle. $l$ is the vehicle length, and $\beta_t$ is the slip angle at the center of gravity of the vehicle. Based on the above formulas, the horizontal velocity $v_x$ and the vertical velocity $v_y$ can be further calculated:

$$\begin{cases} v_{x,t+1} = v_{t+1} \cdot \sin(\beta_t + \psi_t) \\ v_{y,t+1} = v_{t+1} \cdot \cos(\beta_t + \psi_t) \end{cases} \tag{26}$$

Since the velocity is known, the position $s_x$ in the horizontal direction and $s_y$ in the vertical direction can be obtained:

$$\begin{cases} s_{x,t+1} = s_{x,t} + v_{x,t+1} \cdot dt \\ s_{y,t+1} = s_{y,t} + v_{y,t+1} \cdot dt \end{cases} \tag{27}$$

Guided by reward signals, AEV learns how to act in their environment. Learning may be difficult due to sparse or de-layed rewards, so the proper reward is crucial. Considering efficiency and safety, the reward is set as follows:

$$r = \begin{cases} 1 \cdot highest\text{-}speed - 5 \cdot collision, & \text{if not reach the ending point} \\ 1 & , \text{if reach the ending point} \end{cases} \quad (28)$$

where $collision$=1 if AEV collides, otherwise $collision$=0. Similarly, $highest\text{-}speed$=1 if the velocity of the AEV is the highest, otherwise $highest\text{-}speed$=0. This reward function is set to encourage AEV executing a vertical left turn to reach the destination as soon as possible without collision.

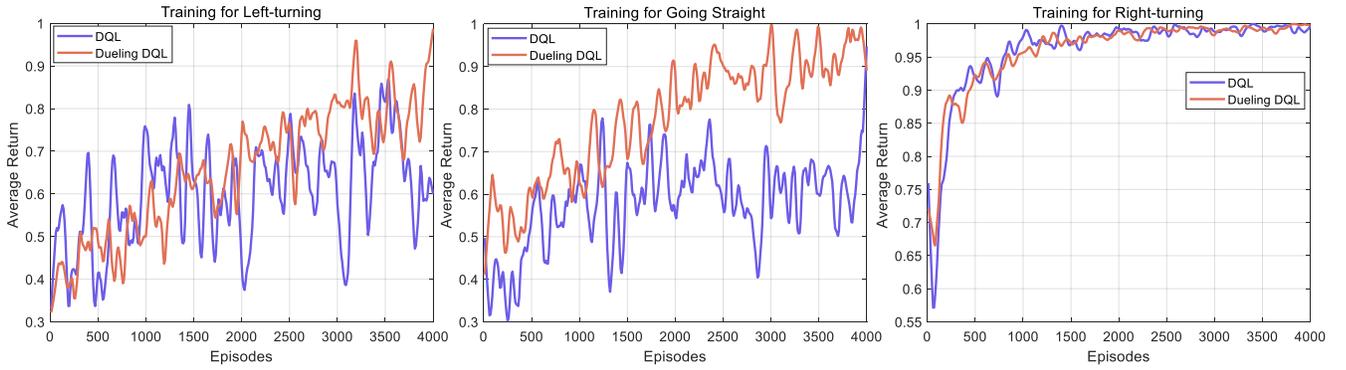

Fig. 5. Average return generated by DQL and dueling DQL in three driving tasks: left-turning, going straight and right-turning.

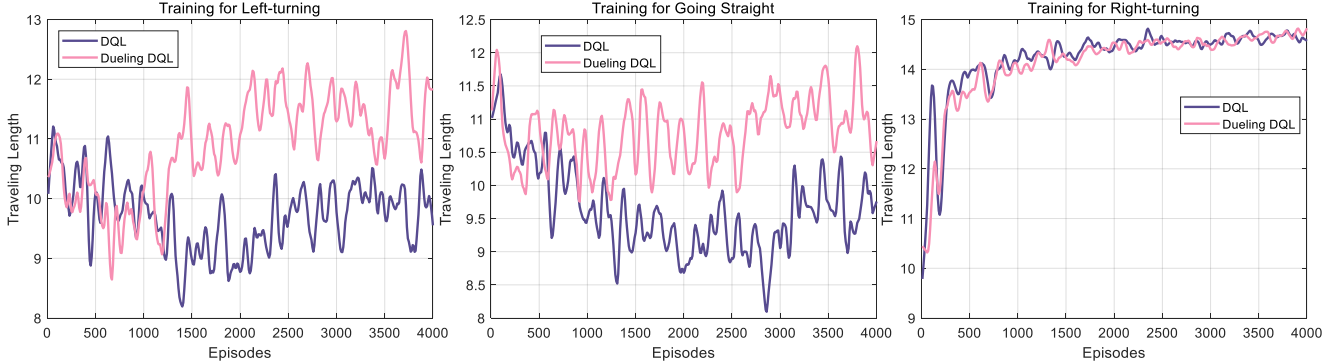

Fig. 6. Compared driving length of the AEV in two DRL approaches.

## IV. RESULTS DISCUSSION AND ANALYZATION

The performance of the proposed decision-making policy is estimated by simulation results in three aspects. First, the training processes of DQL and dueling DQL are compared and illuminated. Then, the trained networks via these two methods from different source driving tasks are transformed into three target tasks, which are left-turning, right-turning, and going straight. The testing reward and success rate are utilized to evaluate the TRL effect. Finally, the dueling DQL with and without TL are compared to verify the effectiveness of TL.

### A. DRL Training for Source Driving Tasks

This subsection describes the training process of DQL and dueling DQL for source driving tasks. As mentioned above, the source driving tasks are left-turning, right-turning, and going straight at the unsignalized intersections. The running objective of the AEV is to drive as fast as possible without collision. For specification, the AEV should promote its speed and avoid crashing the surrounding vehicles. In different driving tasks, the running goal is the same. Hence, the learned knowledge by DRL from one task is able to be transformed into another one.

As depicted in Section II, the intelligent agent in DRL aims to maximize the cumulative reward in (1). Fig. 5 shows the average return (reward) obtained by DQL and dueling DQL in three driving tasks. The number of episodes in these two DRL methods is 4000. In the driving tasks of left-turning and going straight, the reward of DQL has a fluctuant trajectory. It means that it is a little difficult for the AEV to know the driving environment well and choose the appropriate actions. Oppositely, the dueling DQL has an apparent rising trend in the first two tasks. It implies that the agent could acquire a bigger reward in dueling DQL. The relevant decision-making policy is better than DQL. Furthermore, in the right-turning situations, these two techniques are capable of achieving the same return. This is because that the right-turning task is relatively easy. The surrounding vehicles would not often block the AEV, and thus the collision could be easily averted.

To further analyze the control performance of these two approaches, the traveling distances of the AEV are compared in the three driving tasks. This traveling length is affected by vehicle speed, collision condition, and running time. In the left-turning and going straight situations, the traveling length in dueling DQL is more significant than that in DQL. It can be deduced that the vehicle speed is higher in dueling DQL, and the collision rate is lower in this technique. The surrounding vehicles would influence the driving distance of the AEV, too. Since the velocity and position of the surrounding vehicles are randomly defined, the traveling length of the AEV would be affected accordingly. In the right-turning environment, the DQL and dueling DQL are nearly the same. This is caused by traffic conditions, in which the right of way of right-turning is higher. Thus, the AEV could make a right-turning easily without crashing other vehicles.

Finally, as the vehicle speed of the AEV is defined as the state variable in the decision-making problem, the curves of vehicle speed are displayed in Fig. 7. The vehicle velocity of the AEV is mainly determined by the right-of-way. In the intersection environment, the right-of-way indicates which car has the priority to go through the driving scenario. In general, the right-turning and going straight have a higher priority. The left-turning behavior needs to yield other vehicles. This feature in right-of-way could be embodied in the tracks of vehicle speed in Fig. 7. The speed trajectories in going straight and right-turning cases rise with the number of episodes. However, the velocity curve vibrates in the left-turning condition. This is because the random positions of the surrounding vehicles would affect the decisions of the AEV. In other words, the AEV needs to yield other vehicles and catch an opportunity to make left-turning. Moreover, dueling DQL is still better than the DQL in Fig. 7. In conclusion, the dueling DQL has better training performance than DQL.

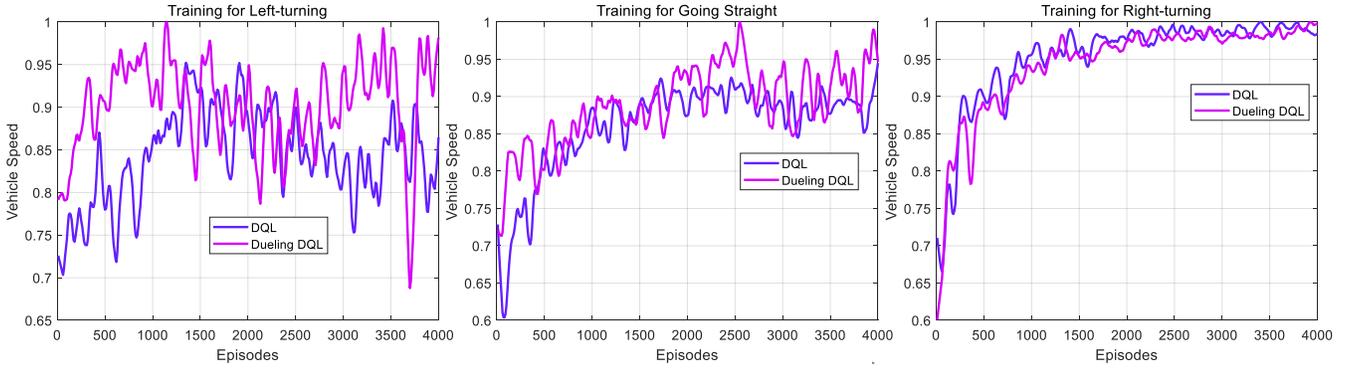

Fig. 7. Vehicle speed of the AEV in the training process of DQL and dueling DQL.

### B. Evaluation of Transferred Knowledge

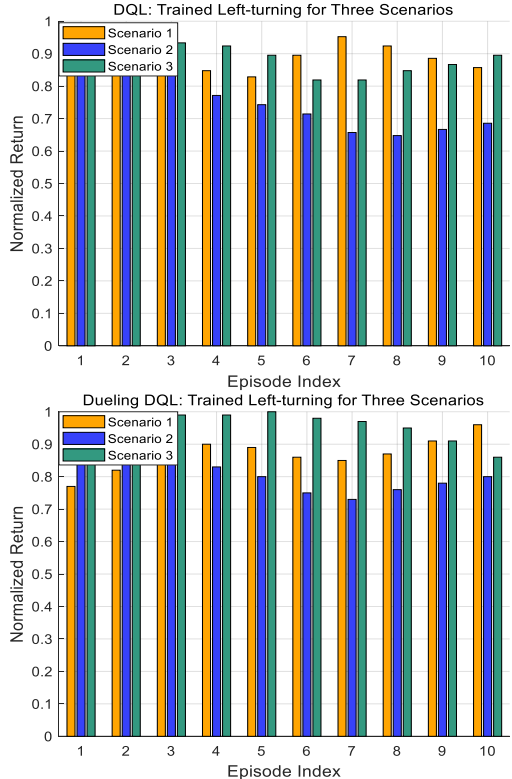

Fig. 8. Reward in the TRL control case: from left-turning to three driving tasks.

The proposed transfer reinforcement learning (TRL)-based decision-making strategy is validated in this subsection. The three driving tasks are indexed as scenario 1, scenario 2 and scenario 3 (sce 1, sce 2 and sce 3 for short) for convenience. The testing thought is arranged as follows: the trained networks in Section IV.A are transformed into the target driving tasks to improve the learning efficiency. Without loss of generality, the source driving tasks are settled as the left-turning and going straight. The target driving tasks are the three scenarios. It implies that the knowledge (trained network) learned from one source task would be applied in three target tasks. The testing episodes are 10.

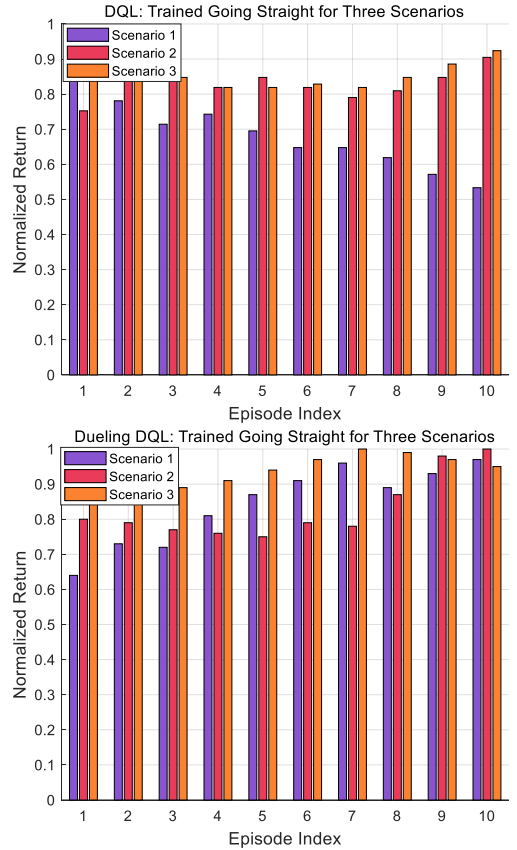

Fig. 9. Normalized return in TRL method: source driving task: going straight, target driving tasks: three driving scenarios.

Fig. 8 describes the related rewards in different target driving scenarios. The source driving task is left-turning. DQL and dueling DQL are also compared in this TRL control framework. From Fig. 8, several appearances can be observed. The rewards in scenarios 1 and 3 are higher than those in scenario 2. Two reasons cause this characteristic. First, the left-turning is the source task, and thus the learned knowledge performs well in the left-turning situation. Second, right-turning is easy enough because it has a higher right-of-way. Furthermore, from the views of different methods, it can be observed that the dueling DQL is better than DQL. It is attributed to the advantage function $A(s, a)$ in the dueling DQL. This function could help the AEV to measure the worth of control action and choose better actions.

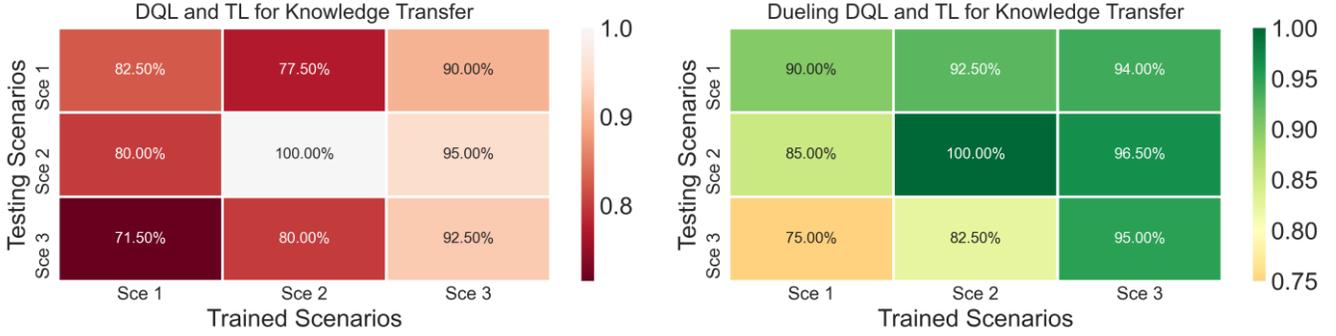

Fig. 10. Success rate of the AEV in two TRL control cases for three driving scenarios.

Then, the source driving task changes to going straight behavior. Fig. 9 depicts the similar distribution of rewards in DQL and dueling DQL. Since the source driving task is the going straight condition, the reward in scenario 2 is higher than that in scenario 1. From Fig. 8 and Fig. 9, it can be concluded that the source driving task would affect the control performance in different driving tasks. Similarly, since the right turning is easy enough, the return in scenario 3 is always large. To compare the two DRL methods in the transferred framework, the two sketches in Fig. 9 are able to be compared. The returns in dueling DQL and TL are usually better than DQL and TL. It indicates that the combination of dueling DQL and TL is a better choice.

Finally, the success rate of the learned knowledge applications is the most concerning factor. The heatmap of the relationship between the source tasks and target tasks is given in Fig. 10. The dueling DQL and TL and DQL and TL are compared and discussed. The number in each cell indicates the success rate of the TRL control cases, which means the collision did not happen. The X-axis means the source driving tasks, and the Y-axis implies the target driving scenarios. In most cases, when the source and target scenarios are the same, the success rate is very high. For example, when these indexes are also 2, the success rate is 100% in these two control cases. This feature conforms to the natural logic in the actual driving environment. By comparing the same cell in DQL and dueling DQL, it can be observed that the dueling DQL and TL case has better effects. It illustrates that the dueling DQL and TL is more suitable for the decision-making problem in this work. The impact of TL will be discussed in the next subsection.

### C. Estimation of TL Effectiveness

The foregoing content describes the training process of the DRL approaches and the performance of the TRL-based decision-making strategies. This subsection focuses on assessing the effects of TL in RL learning procedure. As described in Section II. C, the TL is added into the DRL framework to improve the learning efficiency. With respect to the decision-making problems at intersections, the trained networks from one driving task is transformed into the target mission. Since the dueling DQL is demonstrated to be superior to the DQL in this work, the dueling DQL with and without TL are compared in this part.

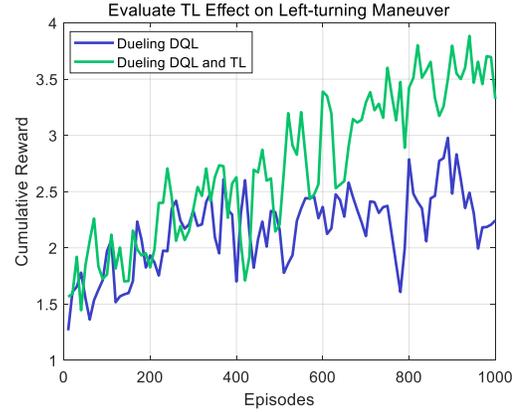

Fig. 11. Cumulative reward dueling DQL with and without TL.

To consider the paper length limitation, the driving task is selected as the left-turning behavior at the unsignalized intersection. The sole dueling DQL, and dueling DQL and TL are compared in 1000 episodes. Fig. 11 shows the accumulative reward in these two methods. Based on the transformed knowledge, the dueling DQL and TL is able to realize higher reward in the training process. It implies that the parameters of the network enable the AEV to understand the driving environment better. As a consequence, the generated decision-making policy is more appropriate for traffic situations. Hence, the TL method could help autonomous vehicles to choose better control actions according to the transformed knowledge.

To display the convergence rate of these two compared approaches, Fig. 12 exhibits the error of Q-table in these two control cases. Four types of error trajectories are given in this figure. From Fig. 12, it is obvious that the initial errors are different. Owing to the learned experiences in dueling DQL and TL, the error of Q-table is small at the beginning. The values of these errors in dueling DQL and TL are lower than those in DQL, which indicates that the evaluated network in the former is closer to the actual one. This feature also explains that the proposed TRL method has a faster convergence rate. Depend-ing on the simulation results in Fig. 11 and 12, it can be discerned that the TL concept is capable of improving the learning rate and guaranteeing the control performance. Since the convergence rate could be promoted, the presented TRL framework is a possible solution to construct the real-time decision-making policy for autonomous vehicles.

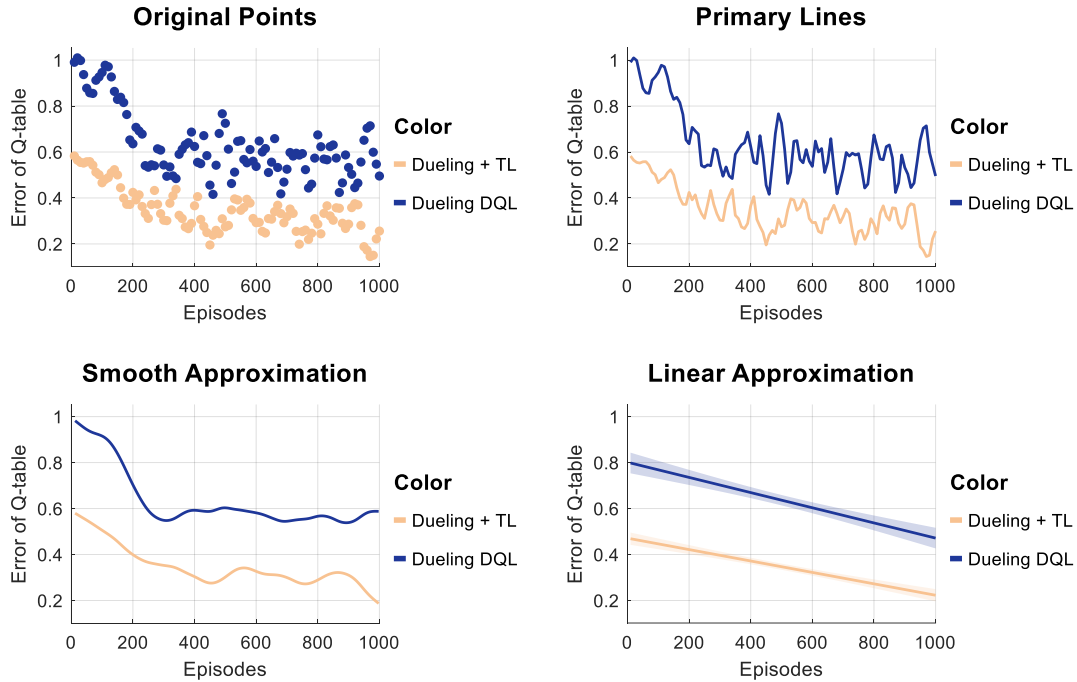

Fig. 12. Convergence rate comparison in the sole dueling DQL and dueling DQL and TL.

## V. CONCLUSION

This paper proposes a transfer reinforcement learning (TRL) framework to improve the control performance and learning efficiency for automated vehicles' decision-making problems. Three driving tasks are involved, which are left-turning, going straight, and right-turning. The learned knowledge from the source driving task is transformed into the target missions. Three aspects of evaluated experiments are conducted to expound the advantages of the combination of dueling DQL and TL. Results indicate that the proposed decision-making policy could improve learning efficiency and performance.

Future works aim to realize the online implementation of the decision-making policies for autonomous vehicles. The connected environment can be discussed. The real-world driving data is able to be applied. More advanced DRL algorithms are capable of being utilized.

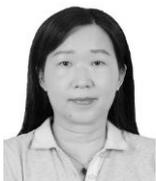

**Hong Shu** was born in Chongqing, China, in 1963. She received the Ph.D. degree in mechanical engineering from the University of Chongqing, Chongqing, China, in 2008. She is currently an associate professor at the School of Automotive Engineering, Chongqing University, Chongqing, China. She has been is a main researcher of over 20 research projects and has published more than 50 academic papers. Her research interests include automated vehicle control, testing and evaluation, electric vehicle dynamics and control, and vehicle eco-driving assistance technology.

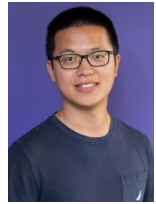

**Teng Liu** (M'2018) received the B.S. degree in mathematics from Beijing Institute of Technology, Beijing, China, 2011. He received his Ph.D. degree in automotive engineering from Beijing Institute of Technology (BIT), Beijing, in 2017. His Ph.D. dissertation, under the supervision of Pro. Fengchun Sun, was entitled "Reinforcement learning-based energy management for hybrid electric vehicles." He worked as a research fellow in Vehicle Intelligence Pioneers Ltd from 2017 to 2018. Dr. Liu worked as a postdoctoral fellow at the Department of Mechanical and Mechatronics Engineering, University of Waterloo, Ontario N2L3G1, Canada from 2018 to 2020. Now, he is a member of IEEE VTS, IEEE ITS, IEEE IES, IEEE TEC, and IEEE/CAA.

Dr. Liu is now a Professor at the Department of Automotive Engineering, Chongqing University, Chongqing 400044, China. He has more than 8 years' research and working experience in renewable vehicle and connected autonomous vehicle. His current research focuses on reinforcement learning (RL)-based energy management in hybrid electric vehicles, RL-based decision making for autonomous vehicles, and CPSS-based parallel driving. He has published over 40 SCI papers and 15 conference papers in these areas. He received the Merit Student of Beijing in 2011, the Teli Xu Scholarship (Highest Honor) of Beijing Institute of Technology in 2015, "Top 10" in 2018 IEEE VTS Motor Vehicle Challenge and sole outstanding winner in 2018 ABB Intelligent Technology Competition. Dr. Liu is a workshop co-chair in 2018 IEEE Intelligent Vehicles Symposium (IV 2018) and has been reviewers in multiple SCI journals, selectively including IEEE Trans. Industrial Electronics, IEEE Trans. on Intelligent Vehicles, IEEE Trans. Intelligent Transportation Systems, IEEE Transactions on Systems, Man, and Cybernetics: Systems, IEEE Transactions on Industrial Informatics, Advances in Mechanical Engineering.

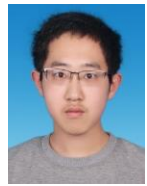

**XingYu Mu** received the B.S degree in Chongqing University, major in Automotive Engineering. He is currently pursuing M.S. degree in Chongqing University, major in Automotive Engineering. His current research focuses on the left-turn decision-making problem of autonomous driving at intersections.

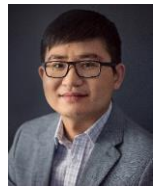

**Dongpu Cao** received the Ph.D. degree from Concordia University, Canada, in 2008. He is currently an Associate Professor and Director of Driver Cognition and Automated Driving (DC-Auto) Lab at University of Waterloo, Canada. His research focuses on vehicle dynamics and control, driver cognition, automated driving and parallel driving, where he has contributed more than 170 publications and 1 US patent. He received the ASME AVTT'2010 Best Paper Award and 2012 SAE Arch T. Colwell Merit Award. Dr. Cao serves as an Associate Editor for IEEE TRANSACTIONS ON VEHICULAR TECHNOLOGY, IEEE TRANSACTIONS ON INTELLIGENT TRANSPORTATION SYSTEMS, IEEE&ASME TRANSACTIONS ON MECHATRONICS, IEEE TRANSACTIONS ON INDUSTRIAL ELECTRONICS and ASME JOURNAL OF DYNAMIC SYSTEMS, MEASUREMENT AND CONTROL. He has been a Guest Editor for VEHICLE SYSTEM DYNAMICS, and IEEE TRANSACTIONS ON SMC: SYSTEMS. He has been serving on the SAE International Vehicle Dynamics Standards Committee and a few ASME, SAE, IEEE technical committees, and serves as the Co-Chair of IEEE ITSS Technical Committee on Cooperative Driving.